\documentclass[conference]{IEEEtran/IEEEtran}
\IEEEoverridecommandlockouts
\usepackage{cite}
\usepackage{amsmath,amssymb,amsfonts}
\usepackage{multirow}
\usepackage{algorithmic}
\usepackage[linesnumbered,ruled]{algorithm2e}
\usepackage{graphicx}
\usepackage{subfig}
\usepackage{textcomp}
\usepackage{mathtools}
\usepackage{flushend}
\usepackage[bookmarks=false]{hyperref}

\def\BibTeX{{\rm B\kern-.05em{\sc i\kern-.025em b}\kern-.08em
    T\kern-.1667em\lower.7ex\hbox{E}\kern-.125emX}}
\begin{document}

\title{VORRT-COLREGs: A Hybrid Velocity Obstacles and RRT Based COLREGs-Compliant Path Planner for Autonomous Surface Vessels}

\author{\IEEEauthorblockN{ Rahul Dubey and Sushil J Louis}
\IEEEauthorblockA{\textit{Computer Science and Engineering, University of Nevada Reno, USA} \\
rdubey018@nevada.unr.edu, sushil@unr.edu}
}

\maketitle

\begin{abstract}
This paper presents VORRT-COLREGs, a hybrid technique that combines velocity obstacles (VO) and rapidly-exploring random trees (RRT) to generate safe trajectories for autonomous surface vessels (ASVs) while following nautical rules of the road. RRT generates a set of way points and the velocity obstacles method ensures safe travel between way points. We also ensure that the actions of ASVs do not violate maritime collision guidelines. Earlier work has used RRT and VO separately to generate paths for ASVs. However, RRT does not handle highly dynamic situations well and and VO seems most suitable as a local path planner. Combining both approaches, VORRT-COLREGs is a global path planner that uses a joint forward simulation to ensure that generated paths remain valid and collision free as the situation changes.
Experiments were conducted in different types of collision scenarios and with different numbers of ASVs. Results show that VORRT-COLREGs generated collision regulations (COLREGs) complaint paths in open ocean scenarios. Furthermore, VORRT-COLREGs successfully generated compliant paths within traffic separation schemes. These results show the applicability of our technique for generating paths for ASVs in different collision scenarios. To the best of our knowledge, this is the first work that combines velocity obstacles and RRT to produce safe and COLREGs complaint path for ASVs. 
\end{abstract}

\begin{IEEEkeywords}
ASV, COLREGs, RRT, Velocity Obstacles.
\end{IEEEkeywords}

\section{Introduction}
In recent years, technological advancements have enabled autonomous surface vessels to operate  beside other autonomous and human operated surface vessels although many challenges still need to be addressed. Safe navigation of ASVs in dynamic environments is one of many such challenges. In maritime navigation, each vessel must follow rules from the International Regulations for Preventing Collision at Sea (commonly known as COLREGs). These rules prescribe different types of maneuvers that a vessel must perform under different types of collision scenarios to avoid any unfortunate incidents.
Generating a safe and COLREGs complaint path for an ASV to follow between a starting and a goal location is a difficult task in dynamic environments because of changes in the environment. The problem becomes even more complex when dealing with large numbers and types of vessels in close vicinity where each vessel follows COLREGs. Changes made in the course and/or speed of any ASV might create a ripple effect that forces other ASVs to take appropriate actions and may quickly lead to large changes in the state of environment.
In the past, a variety of approaches have been proposed that obey COLREGs, such as rapidly random tree~\cite{RRT}, velocity obstacles~\cite{kuwata2013safe}, fuzzy logic, model predictive control (MPC)~\cite{MPC}, artificial potential fields (APF)~\cite{APF}. However, these approaches do not scale well to large numbers of vessels and multiple COLREGs rules. Additionally, some of these approaches, such as RTT, have a high computational complexity. 

In this paper, we present a technique, VORRT-COLREGs, that combines RRT and VO to generate safe and COLREGs complaint paths for ASVs in different types of collision scenarios. RRT samples a 2D location for an ASV to move to and VO checks whether moving to the new location is safe and whether the actions of the ASV will be in accordance with COLREGs. If the newly generated sample is safe and COLREGs complaint then this location is added to a list of way points otherwise the location is discarded. Note that due to changes in the environmental state that includes vessels position, speed, and course, a previously COLREGs complaint path may becomee invalidated and thus a new path needs to be found. To avoid this, we employ joint forward simulation (JFS) to compute the future locations of all vessels assuming that the speed and course of vessels remains the same. When a path becomes invalid, we compute a new path. JFS is a popular way of predicting the future locations of autonomous agents and have been extensively used in robotics.

Earlier work~\cite{RRT} describes COLREGs complaint path generation using virtual obstacles with RRT in the open ocean. We extend this work and use VORRT-COLREGs to generates safe and COLREGs complaint paths for many different collision situations including Traffic Separation Schemes (TSS). To the best of our knowledge, this is the first work that combines RRT and VO to generate path for ASVs in multiple different scenarios.

Experiments were conducted with different numbers of ASVs and under different types of collision scenarios. We started with simple collision situations between two ASVs: crossing, head-on, and overtaking. Results show that VORRT-COLREGs generated safe and COLREGs complaint path under each of the three scenarios. Next, we considered more than two vessels where multiple types of collision occurs simultaneously and results show that our technique generated compliant COLREGs paths. Lastly, we considered a TSS where multiple vessels move within a narrow channel and an ASV crosses the channel. Results show that VORRT-COLREGs generated paths that safely navigated the ASV from one side to the other side of the channel.

The remainder of this paper is organized as follows. Section~\ref{RelatedWork} outlines related work in the area of autonomous surface vessels navigation. Section~\ref{Simulation Environment} describes collision regulations.
In Section~\ref{Methodology} we describe our proposed COLREGs complaint path generation algorithm. Section~\ref{Results} presents results and discussion. Finally, the last section draws conclusions from our results and discusses future work.

\section{Related Work} {\label{RelatedWork}}

Reports indicate that more than $80$ percent of maritime collision accidents are caused by human error~\cite{data}. Autonomous navigation of maritime surface vessel may help to reduce collision accidents by efficient and collision free route planning. Generating safe paths that are in compliance with collision regulations is difficult because of the dynamic nature of the operating environment. Vessels' positions, courses, speeds, sizes, and other geographical restrictions affect path planning. Any mistake made by any vessel operator may quickly result in scenarios where collision between vessels is inevitable. For example, a route planned for an ASV may cause other vessels to alter their course or speed while following COLREG guidelines but they may not have enough time or space to maneuver to avoid colliding with still other vessels in a scenario. 

We investigate collision free path planning navigation for ASVs in compliance with collision regulations at sea. Prior work has used techniques based on RRT~\cite{RRT}, velocity obstacles~\cite{kuwata2013safe}, and model predictive control~\cite{MPC} to control the movement of ASVs while avoiding obstacles and following COLREGs. However, these approaches may not scale to heavy traffic environments such as those found in Traffic Separation Schemes (TSS). To the best of our knowledge, VORRT-COLREGs is the first approach to combine velocity obstacles with RRT to quickly find a COLREGs complaint collision free path to a destination. 

Researchers have used different techniques to generate COLREGs complaint paths. In~\cite{APF}, artificial potential fields in combination with a line-of-sight based algorithm was introduced. This does not incorporate the large mass and inertia of vessels, hence vessels may fail to come to and follow a desired course in time to avoid collision. In addition, there is no handling of maintaining a safe minimum distance from other ships. Johansen~\cite{model-predictive-control} proposed model predictive control and used used simulated predictions of trajectories of obstacles and other vessels. Loe~\cite{dynamic-window-based} used a dynamic window based method for ASV navigation. However, in both approaches, other vessels in their test scenarios are non COLREG complaint and have random or fixed velocities. We address all these issues using our proposed technique.

Pranay~\cite{A-star} used A{*} under dynamic environment conditions to generate paths for ASVs. An iterative 
A{*} heuristics was used to find a shortest path while maintaining adequate distance from colliding vessels on $15$ different environments where autonomous vehicle moves at speed of $16.8$ m/s and the target location moves at $1.7$ m/s.
In~\cite{shah2016resolution} Shah presented an Adaptive Risk and Contingency-Aware Planner (ARCAP) to generate paths in a lattice. Shen~\cite{shen2019automatic} used a Deep Q Network (DQN) for multi ship collision avoidance while following rules of the road. Li~\cite{li2019optimizing} proposed rolling horizontal optimization approach to minimize course alteration and time efficiency while maneuvering vessels. Synchronous Branch and Bounce (SyncBB), Dynamic Programming Optimization Protocol (DPOP), and Asynchronous Forward Bounding (AFB) was used for multi ship avoidance in ~\cite{li2019distributed}. Kim~\cite{kim2017distributed} presented Dynamic Stochastic Search Algorithm (DSSA) and
Chen~\cite{chen2018distributed} introduced Alternative Direction Method of Multiplier (ADMM) for ship control and the
room-to-maneuver algorithm, which contains a safe velocity
space for multiple ships, was proposed by Degre~\cite{degre1981collision}. 

Researchers have also formulated this problem as an optimization problem and used different optimization techniques to optimize path. Liang in \cite{multi-objective}, proposed a multi objective particle swarm optimization based on-line route planning framework for ASVs that minimizes collision and ensure COLGREGs compliance. The framework re-plans the route in dynamic environment and scales with multiple target vessels. Shengke~\cite{ni2018modelling} use genetic algorithms to optimize paths that avoid collisions at open sea. Jun~\cite{Multi_objective_collision} proposed a modified fuzzy dynamic risk of collision model based on time and space collision risk index to generate paths and used multi objective genetic algorithm to optimize paths based on time, economy, COLREGs, and safety. 
Compared to the above approaches, VORRT-COLREGs provides fast, safe, near-optimal, physically realizable paths for ASVs. 

\section{COLREGs}{\label{Simulation Environment}}

\begin{figure*}
    \centering
    \includegraphics[scale= 0.55]{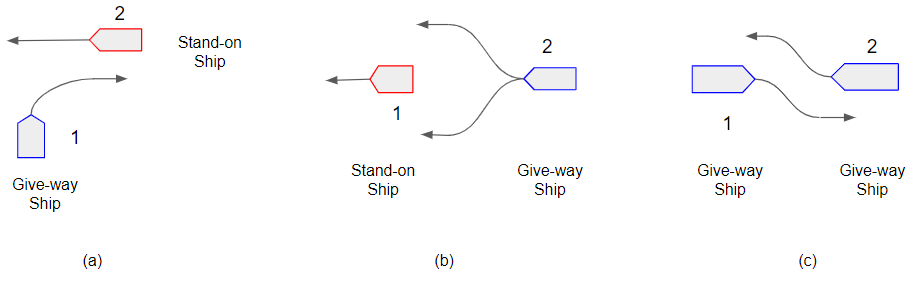}
    \caption{COLREGs guidelines for (a) crossing, (b) over-take, and (c) head-on.}
    \label{cross-overtake}
\end{figure*}

Collision regulation guidelines prescribe actions that different vessels must perform when encountering a potential risk of collision. Figure ~\ref{cross-overtake}, shows three different scenarios with potential risk of collision. In this figure (a) is a crossing scenario where $1$ is the give-way vessel and $2$ is the stand-on vessel, (b) is an over-taking scenario where $1$ is the stand-on vessel and $2$ is the give-way vessel, and (c) is a head-on scenario where both vessels must give-way.

\subsection{Risk of Collision}

In this paper, when the distance between our vessel (ownship) and a target vessel is less than a Minimum Threshold Distance (MTD), we check for potential risk of collision. If the distance at the Closest point of Approach (CPA) is below this MTD we assume that there is a potential risk of collision and use VORRT-COLREGs to generate a path for ownship to follow.

\subsection{Rule-Based Encounter Type Classification}

If there is a risk of collision and ownship is the give-way vessel, the scenario or encounter type is classified according to collision regulation rules. In this work we consider only rules $13$ to $17$ which dictate the actions of power driven vessels to avoid collision. These rules define five different types of collision situations: head-on, starboard crossing, port crossing, overtaking, and being overtaken. In case of head-on and crossing from starboard, the give way vessel must change course to starboard. When being overtaken or when you are the stand on vessel in a crossing situation, you should maintain speed and course. Finally, when overtaking, ownship can overtake on the port or starboard sides of the target. Here are pertinent rules from COLREGs that VORRT-COLREGs follows:

\begin{itemize}
    \item Overtaking (Rule 13(b)): A vessel is coming up with
another vessel from a direction more than 22.5 degrees
abaft her beam.

\item Head-on (Rule 14(a)): Two power-driven vessels are
meeting on reciprocal or nearly reciprocal courses so as
to involve risk of collision.

\item Crossing (Rule 15): Two power-driven vessels are crossing so as to involve risk of collision.

\item Action by give-way vessel (Rule 16): Every vessel which
is directed to keep out of another vessel shall, so far as
possible, take early and substantial action to keep well
clear.

\item Action by stand-on vessel (Rule 17(a)(i)): Where one of
two vessels is to keep out of the way the other shall keep
her course and speed.
\end{itemize}

\section{Methodology}{\label{Methodology}}

We propose a hybrid of VO and RRT to generate COLREGs compliance paths for ASVs. We start by describing velocity obstacles followed by RRT. Later, we combine VO and RRT and describe VORRT-COLREGs to generate COLREGs complaint paths for ASVs.

\subsection{Velocity Obstacle }

\begin{figure}
    \centering
    \includegraphics[scale= 0.35]{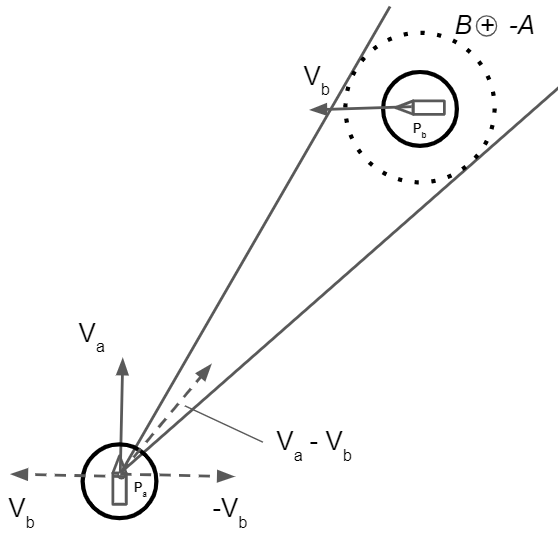}
    \caption{Velocity obstacle cone of A and B.}
    \label{VO_Fig}
\end{figure}

Velocity obstacles~\cite{van2008reciprocal} was proposed in 2008 as a new concept for real-time multi agent navigation. Let $A$ and $B$ be two vessels positioned at $P_{a}$ and $P_{b}$ and moving in the direction of their arrows with velocities of $V_{a}$ and $V_{b}$ as shown in Figure~\ref{VO_Fig}. We assume that $A$ is an ASV and $B$ is a human controlled vessel. The velocity obstacle $VO_{b}^{a}(\Vec{V_{b}})$ of vessel $B$ to the ASV is the set consisting of all those velocities $V_{a}$ for ASV that will result in a collision at some moment in time with vessel $B$ moving at velocity $V_{b}$. Figure~\ref{VO_Fig} shows the velocity obstacle $VO_{b}^{a}(\Vec{V_{b}})$ of a disc-shaped vessel $B$ to a disc-shaped ASV. For simplicity we consider disc-shaped vessels where the radius of the enclosing disc is the length of the vessel. In order to avoid collision, two vessels' representative discs must not touch each other at any given time. Let $A \oplus B $ denote the Minkowski sum of vessels $A$ and $B$ and let$-A$ denote the vessel $A$ reflected in its reference point.
\[A \oplus B  = \{a + b | a \in A, b \in B \}, -A = \{-a | a \in A\}\]

$ \lambda(\Vec P, \Vec V) $ denotes a ray staring at $P$ and heading in the direction of $V$ as shown in equation~\ref{VOeq1}. If the ray starts at $P_{a}$, heading in the direction of the relative velocity of $A$ and $B$ and intersects the Minkowski sum of $B$ and $-A$ centered at $P_{b}$, then the velocity $V_{a}$ is in the velocity obstacle of $B$. Equation~\ref{VOeq2} calculates velocity obstacle $VO_{b}^{a}(\Vec{V_{b}})$ of vessel $B$ to the ASV $A$.

Figure~\ref{VO_Fig} shows a cone that represents $VO_{b}^{a}(\Vec{V_{b}})$ and the relative velocity ($V_{a} - V_{b}$) lies inside this cone. This means that if the ASV keeps its heading and speed (velocity), there will be a future collision. 
As per rule $15$, this is a crossing scenario where the ASV (ownship) is the give way vessel and the other vessel is the stand on vessel. Rule $15$ also says that if the circumstances of the case admit, avoid crossing ahead of the other vessel. 
Thus, the ASV needs to change its course to starboard to avoid collision in accordance with the nautical rules of the road. 
Furthermore, in the open ocean, vessels may try to keep miles away from each other. The minimum distance between two moving vessels is the distance at closest point of approach (CPA). Equation~\ref{tCPA} computes time to CPA and equation~\ref{dCPA} computes the distance at CPA between the ASV and other vessels.
\begin{equation}
    \lambda(\Vec P, \Vec V) = \{\Vec P + t\Vec V | t \geq 0\} 
    \label{VOeq1}
\end{equation}

\begin{equation}
    VO_{b}^{a}(\Vec{V_{b}}) = \vec{V_{a}}|\lambda(\Vec{P_{a}}, \Vec{V_{a}} - \Vec{V_{b}})
    \label{VOeq2}
\end{equation}
\begin{equation}
  t_{cpa} =
    \begin{cases}
      \frac{(\Vec{P_{a}}  - \Vec{P_{b}}).(\Vec{V_{a}}  - \Vec{V_{b}})}{\lVert(\Vec{V_{a}}  - \Vec{V_{b}})\rVert^2 } & \text{if $\lVert(\Vec{V_{a}}  - \Vec{V_{b}})\rVert > 0$}\\
      0 & \text{otherwise}
    \end{cases}
    \label{tCPA}
\end{equation}
\begin{equation}
  d_{cpa} =
  \lVert(\vec{P_{a}} + t_{cpa}*\vec{V_{a}}) - (\vec{P_{b}} + t_{cpa}*\vec{V_{b}})\rVert 
    \label{dCPA}
\end{equation}
We thus want the ASV to change course to not only avoid collision but to keep at-least a minimum distance from the other vessel as well. In Figure~\ref{VO_Fig} VO provides a safe region for the ASV, rule $15$ prescribes turning to starboard, and the minimum CPA distance constraint further restricts part of the safe region on the starboard side provided by VO. We describe the RRT algorithm and how we utilize this information and constraints during the RRT tree growth phase.

    
\subsection{Rapidly Exploring Random Tree (RRT)}
RRT~\cite{lavalle1998rapidly} was introduced by LaValle in $1998$ as a path planning algorithm. In $2018$, Chiang~\cite{RRT} introduced COLREG-RRT that generated paths for ASV in various collision situations and with different numbers of vessels and our work is closely related to this paper. COLREG-RRT generates different virtual obstacles in different types of collision scenarios and grows RRT trees while avoiding virtual obstacles. However, COLREG-RRT does not guarantee that generated paths will always follow rules of the road, does not explicitly handle overtaking situations, and there is no minimum distance constraint to ensure that vessels do not come dangerously close to each other.

\begin{algorithm}
    \SetKwInOut{Input}{Input}
    \SetKwInOut{Output}{Output}
    
    \Input{$\mathcal{S} = (s_{a}, s_{1}, s_{2}, ..., s_{m})$ }
    \Output{$\mathcal{P}$}
    $\mathcal{N}$ = []\;
    RootNode = Initialize RRT$(s_{a}, s_{1}, s_{2}, ..., s_{m})$\;
    $\mathcal{N}.append(RootNode, 0)$\;
    $\mathcal{N}$, Goal = VORRTTree($\mathcal{N}$, Goal=1,
    itr, $\mathcal{S}$)\;
    \If{!Goal}{
      $\mathcal{N}$, Goal = VORRTTree($\mathcal{N}$, Goal=0,
    itr, $\mathcal{S}$)\;
       }
       $\mathcal{P}$ = getPath($\mathcal{N}$)\;
    return($\mathcal{P}$)\;
    \caption{VORRT-COLREG}
    \label{VORRT_Algo}
\end{algorithm}
\subsection{VORRT-COLREGs}
We present a hybrid of VO and RRT that generates safe paths in compliance with COLREGs. VORRT-COLREGs works in all types of collision situations and is given by algorithm~\ref{VORRT_Algo}. The algorithm takes the environmental states, $\mathcal{S} = (s_{a}, s_{1}, s_{2}, ..., s_{m})$ and returns a path, $\mathcal{P}$, for ASVs to follow. Here $s_{a}$ refers to the current state of an ASV that includes ASV's position, heading, and speed. $s_{1}, ..., s_{m}$ refer to the states of other vessels.
The algorithm begins with an empty list of node $\mathcal{N}$. RRT initializes a root node where the node contains state information of all vessels with timestamp. Since RRT generates new nodes randomly, it may take a longer time to find a path between the current location of the ASV and its goal location. Thus, we introduce a bias by setting a parameter to generating new nodes in the direction of the goal location from ASV's current location. {\tt VORRTTree} in algorithm~\ref{VORRT_Algo} grows the tree and takes four parameters; $\mathcal{N}$, moving towards goal (true/false), maximum number of iterations, and the environmental state. {\tt VORRTTree} grows the tree for $200$ steps and if it can reach the goal location with $200$ steps, extracts and returns the path from the list of nodes ($\mathcal{N}$). If the goal was not reached then the tree grows for another $200$ steps without bias towards goal but following the nautical rules of the road. At the end of this second $200$ steps, if the goal was reached, the algorithm extract and returns the path from the list of nodes. If goal was not reached then we run VORRT-COLREGs again or conclude that no safe path exists.

\begin{algorithm}
\caption{VORRTTree Growth}
\label{VORRTTree_Algo}
    \SetKwInOut{Input}{Input}
    \SetKwInOut{Output}{Output}
    
    \Input{$\mathcal{N}$, Goal, itr, $\mathcal{S}$ }
    \Output{$\mathcal{N}$, Goal}
    
    \While{i $\leq$ itr}{
      \eIf{$Goal$}{
      $pos  = getGoal()$\;}
      {$pos  = randomSampling()$\;}
  $node_{selected} = getClosestNode(\mathcal{N}, pos)$\;
  $pos_{new}, \mathcal{S'} = JFS(node_{selected},pos)$\;
  inCollision = $VOCollision$($pos_{new}$, $\mathcal{S'}$)\;
  \eIf{inCollision}{
  $/*Discard*/$\;
  }
  {
  $node_{new} = getNode(pos_{new}, \mathcal{S'})$\;
  $\mathcal{N}.append(node_{new}, node_{selected})$\;
  i++\;
  }
  }
    Goal = reachedGoal($node_{new}$)\\
    return $\mathcal{N}$, Goal
\end{algorithm}
The VORRTTree growth algorithm is given by Algorithm~\ref{VORRTTree_Algo}. The algorithm works in a similar fashion to vanilla RRT and the only difference is that when a position/location is sampled in two dimensional space, we use a joint forward simulation to compute future states of our ASV and other vessels. The JFS takes the closest node ($node_{selected}$) from this sampled position and computes future states of each vessel. Remember each node contains each vessel's state information (position, speed, and heading) at a given time and thus an ASV with position, speed, and heading identified by $node_{selected}$ moves in the direction of the RRT sampled position for a distance of $d'$ meters. In this paper, $d'=100$ meters as this works well for our vessel speeds and sizes. Other vessels move in the direction of their respective goal locations with constant speed and heading. After travelling $d'$ distance each vessel reaches different locations (state, $\mathcal{S'}$) and we check whether our ASV is at a safe (non VO-colliding) location using velocity obstacles. If the new ASV location is not safe then we discard this new location and  RRT sample a new point/location.

Velocity obstacle based collision checking is shown by Algorithm~\ref{VOCollision}. The algorithm iterates over each vessel ($\mathcal{S'} = {s_{a}, s_{1},..,s_{m}}$) and computes CPA distance between the ASV and other vessels. It also computes whether ASV position ($pos_{new}$) lies in the velocity cone between the ASV and other vessels. If $pos_{new}$ lies inside velocity cone or CPA distance is less than a threshold distance ($d_{th}$) then either there is a risk of collision or collision has occurred. Moving to $pos_{new}$ is not safe and must be discarded.
\begin{algorithm}
    \SetKwInOut{Input}{Input}
    \SetKwInOut{Output}{Output}
    
    \Input{$pos_{new}$, $n_{selected}$, $\mathcal{S'}$}
    \Output{collision}
    collision = false\\
    $\vec{P_{a}} = s_{a}.pos$\\
    $\vec{V_{a}} = s_{a}.vel$\\
    \For{s in $\mathcal{S'}$}{
      $\vec{P_{v}} = s.pos$\\
      $\vec{V_{v}} = s.vel$\\
      $d_{CPA} = findCPADistance(\vec{P_{a}}, \vec{P_{v}}, \vec{V_{a}}, \vec{V_{v}}$)\\
      \If{$pos_{new}$ in $\vec{V_{a}}|\lambda(\Vec{P_{a}}, \Vec{V_{a}} - \Vec{V_{v}}$) $|| d_{th} \leq d_{CPA}$ }{
      \eIf{$\vec{P_{a}} == \vec{P_{v}}$}{
      Do nothing (vessel is the ASV)\\
      }{
      collision = true \\
      }
      }
  }
    return collision\;
    \caption{Velocity Obstacle Collision Checking}
    \label{VOCollision}
\end{algorithm}

\subsection{Joint Forward Simulator (JFS)}
In order to generate safe and COLREGs complaint path, we use a joint forward simulator to check whether each node generated during VORRT tree growth is safe. Each node in the VORRT tree keeps position, speed, and heading of each vessel. Using these information JFS computes the potential for collision between the ASV and other vessels for each newly generated node in the VORRT tree. If for the given node, a collision is detected then that node will be not be added to the VORRT tree. Otherwise the node will be added in the tree and JFS will compute the position, speed, and heading for each vessel for the last added node in the tree. JFS repeats these computations at each time step. It should be noted that modern ocean-going ships are required to be equipped with an automatic identification system which broadcasts the current course, speed and destination~\cite{campbell2012review}. Therefore, we assume this information is available to the JFS. In this preliminary work we only consider rules $13$ to $17$ in ~\cite{guard2012navigation}.
We are specifically interested in situations where ownship is the give way vessel.

\begin{figure}
    \centering
   \includegraphics[width=0.40\textwidth,height=5cm]{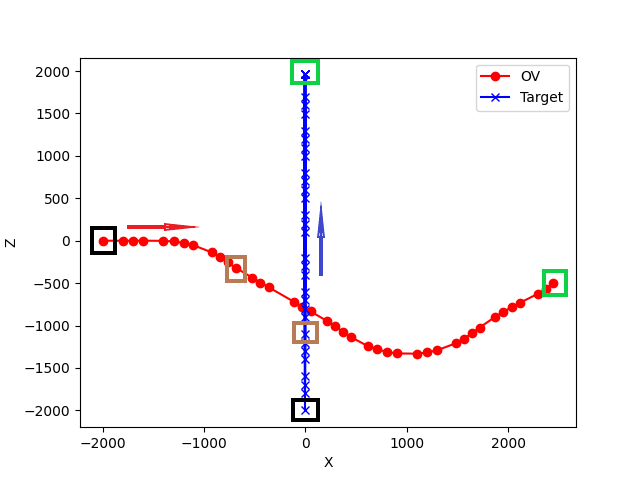}
    \caption{ Crossing situation where the ASV is the give way and target vessel is a stand on vessel. The red line show the path generated by VORRT-COLREGs algorithm that avoids collision while following rules.}
    \label{fig:crossing}
\end{figure}
\begin{figure*}
    \centering
    \subfloat[Overtaking]{\includegraphics[scale=0.4]{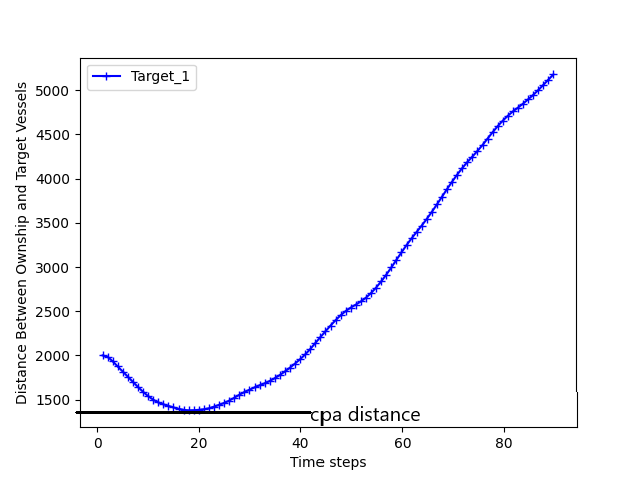}}
    \subfloat[Head-on]{\includegraphics[scale=0.4]{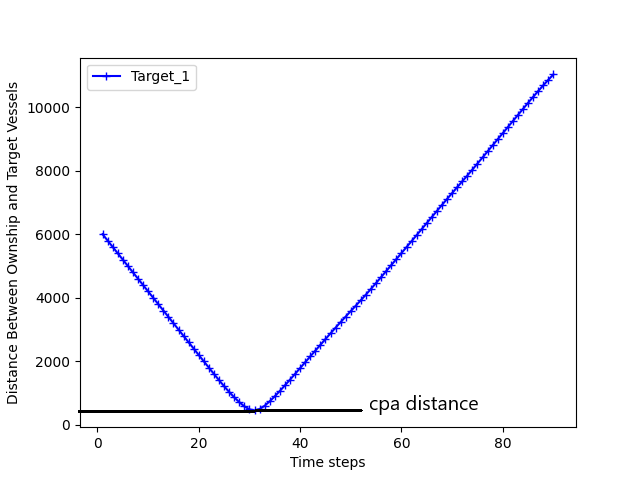}}
    \caption{Distance between the ASV and the target vessel at different time steps. (a) ASV is overtaking the target vessel, (b) when both vessels are head-on.}
    \label{fig:Distance_single_collision}
\end{figure*}
\begin{figure*}
    \centering
    \subfloat[Four Vessels]{\includegraphics[scale=0.33]{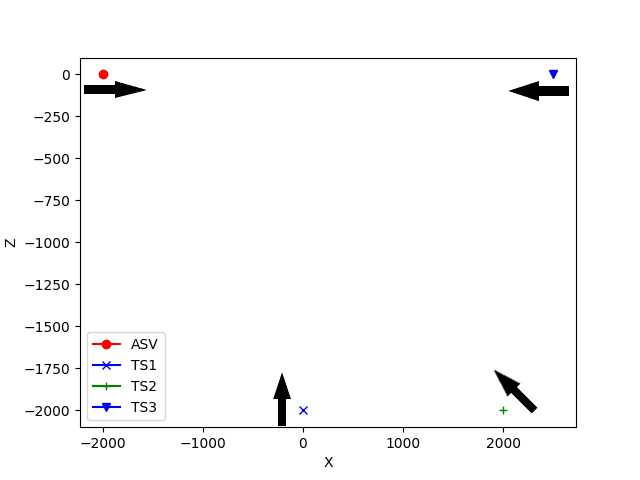}}
    \subfloat[Locations of vessels Vessels]{\includegraphics[scale=0.33]{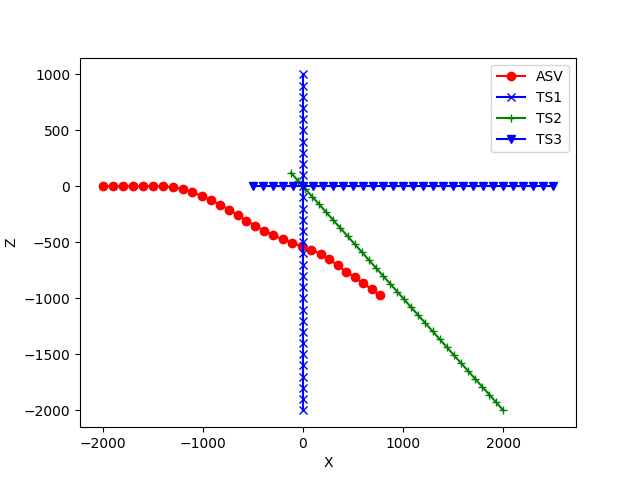}}
    \subfloat[CPA Distances]{\includegraphics[scale=0.33]{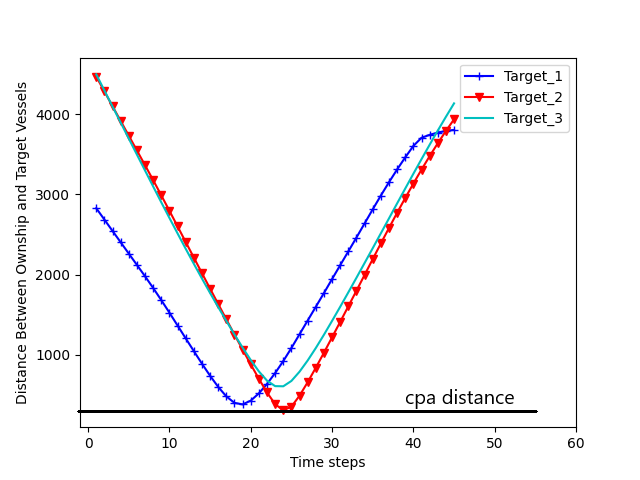}}
    \caption{Multi vessel collision situation. (a) starting location of four vessels moving in the direction of arrow, (b) shows patg travelled by vessels, and
    (c) Distance between the ASV and three target vessels at different time steps.}
    \label{fig:Distance_multi_vessel}
\end{figure*}

\begin{figure*}
    \centering
    \subfloat[A complete path crossing TSS]{\includegraphics[scale=0.5]{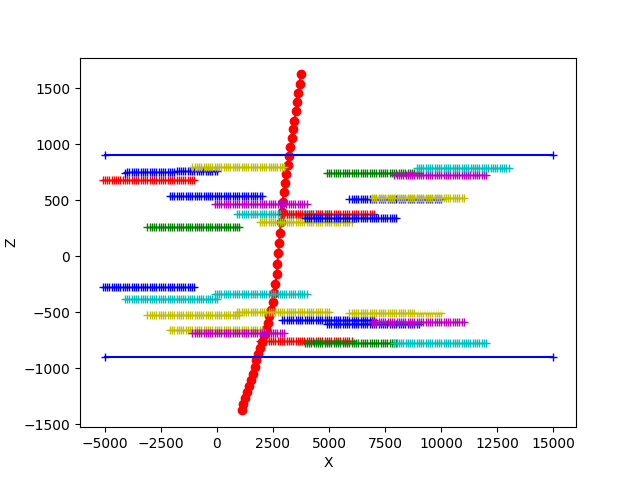}}
    \subfloat[TSS entrance, $d_{min} =  135$ meters]{\includegraphics[scale=0.5]{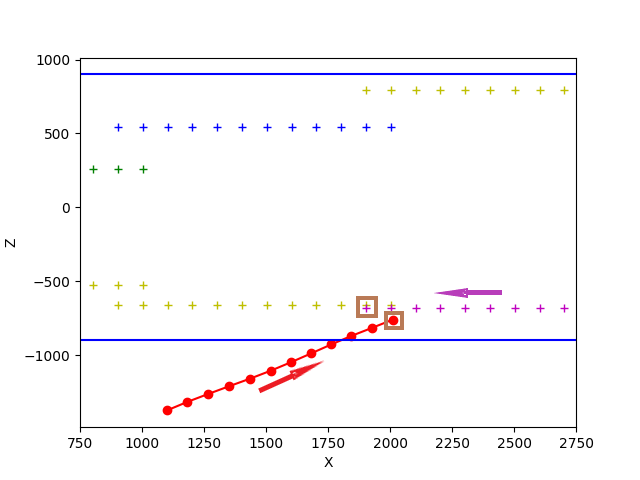}}\\
    \subfloat[Inside TSS,  $d_{min} =  350$ meters]{\includegraphics[scale=0.5]{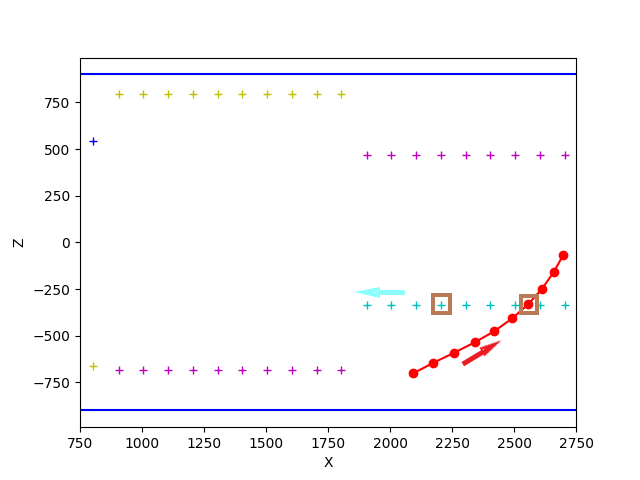}}
    \subfloat[Leaving TSS, $d_{min} =  319$ meters]{\includegraphics[scale=0.5]{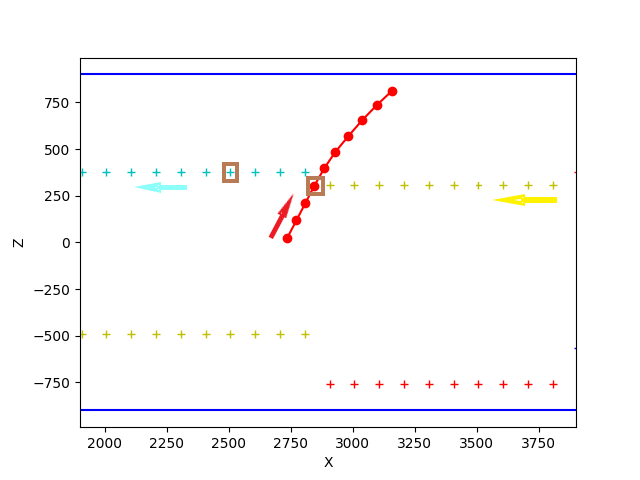}}
    \caption{VORRT-COLREGs generated a safe path for ownship to cross a TSS. Traffic in the TSS is moving from right to left where TSS is one nautical mile in width.  In figures (b), (c), and (d) $d_{min}$ is cpa distance and brown boxes are the closest points of approach.} 
    \label{fig:TSS}
\end{figure*}
\section{Results and Discussion}{\label{Results}}
We conducted experiments with single vessel and multiple vessels under different collision scenarios. 
\subsection{Single Vessel Collision}
We conducted experiments on three baseline test collision scenarios with two vessel. Figure~\ref{fig:crossing} shows
two vessels; ownship (OV)/ASV that is controlled by VORRT-COLREGs and a target vessel in a crossing situation. In this scenario as per rules, ownship is the give way vessel, the target vessel is the stand on vessel, and ownship must turn to starboard to avoid collision. In the figure, black boxes represent starting locations of vessels, brown boxes represent the location of vessels at an intermediate time step, and green boxes represent goal locations of vessels. The figure shows that ownship plots and follows a smooth, physically realizable,  COLREGs compliant, collision avoiding path to the goal.

Figure~\ref{fig:Distance_single_collision}(a) shows the distance between the ASV and the target vessel at different time steps in the simulation, where the ASV is overtaking the target vessel.
The minimum distance, that is, distance at closest point of approach between the two vessels is $1365$ meters as shown by the black line. Similarly, Figure~\ref{fig:Distance_single_collision}(b) shows distance between the ASV and the target vessel when both vessels are head-on to each other at different time steps. Here, the CPA distance was found to be $470$ meters. For the experiments, we set minimum CPA distance threshold to $200$ meters. This minimum CPA distance threshold is a parameter to VORRT-COLREGs and can be easily changed to reflect a Captain's preference or to reflect the scenarios. For example, the minimum CPA distance threshold may need to be set much lower inside a traffic separation scheme where even large vessels may routinely come within several hundred yards of each other.

\subsection{Multi Vessel Collision}
These results on the above simple test scenarios indicate that VORRT-COLREGs generates viable paths for collision scenarios with two vessels. To increase the complexity of the problem we considered a scenario where there is potential risk of collision with multiple vessels. Figure~\ref{fig:Distance_multi_vessel}(a) shows the starting location of an ASV and three target vessels moving in the direction of the arrows. The ASV controlled by VORRT-COLREGs is the give way vessel in both a crossing situation and a head-on situation. VORRT-COLREGs generated a path for the ASV that resulted in safe navigation and the minimum CPA distance is $300$ meters with the target vessel two ($TS2$) and greater for the other vessels. 

\subsection{Traffic Separation Scheme}
We also conducted experiments in scenarios where an ASV/ownship crosses a traffic separation scheme as shown in  Figure~\ref{fig:TSS}. We created a one nautical mile wide channel where normal traffic is moving from right to left and an ASV aims to cross the channel while following nautical rules of the road. Figure~\ref{fig:TSS}(a) shows a complete path generated by VORRT-COLREGs. Zooming in at different time steps we observe the positions of the ASV, other vessels in the TSS and distance between vessels.

Figure~\ref{fig:TSS}(b), (c), and (d) show locations of vessels zoomed in at different time steps. The minimum CPA distance between ASV and all target vessels is $135$ meters. Since, the width is small and vessels do not have a large space to maneuver, we reduced the safe distance between vessels from $200$ to $100$ meters. Note that the COLREG-RRT algorithm in~\cite{RRT} would have been unable to generate viable paths for an ASV to cross this TSS because of the continuous traffic movement in the channel. To generate COLREGs complaint paths, COLREG-RRT assumes a virtual obstacle where one end of the obstacle is in front of colliding vessel and the other end is far away in the direction of vessel movement. Since traffic is always moving in the TSS, this results in the  creation of virtual objects placed in the channel at all times. This does not allow COLREG-RRT to successfully generate paths in a TSS.     

We have shown that VORRT-COLREGs generates collision avoiding paths in accordance with COLREGs for baseline two-vessel and heavy traffic TSS scenarios. We believe that this provides preliminary evidence for the potential of our approach for safe autonomous ship driving in accordance with the nautical rules of the road.  

\section{Conclusions}{\label{Conclusions}}
In this paper, we presented VORRT-COLREGs to generate safe and COLREGs complaint paths for ASVs. We combine rapidly epxloring random trees and velocity obstacles to generate paths where RRT generate samples in the space and velocity obstacle check whether moving in the direction of the newly generated sample is safe and in accordance with COLREGs. Experiments on different scenarios, with different numbers of vessels have shown that our technique successfully generated viable safe paths. Results on baseline two vessel and multiple vessel scenarios show the effectiveness of our technique. We also conducted experiments in a traffic separation scheme with $15$+ vessels and show that VORRT-COLREGs generates safe paths. To the best of authors knowledge this is the first work that combines VO and RRT to generate COLREGs complaint paths for ASVs.

In this future, we plan to extend this work by creating an automated scenario generator and testing the algorithm on hundreds of generated scenarios with differing numbers and densities of vessels in the open ocean and in traffic separation schemes. This will enable scaling of the approach and characterization of the limits of the approach. 

\section*{Acknowledgments}
This work was supported by grant number N00014-17-
1-2558 from the Office of Naval Research. Any
opinions, findings, and conclusions or recommendations expressed in this
material are those of the author(s) and do not necessarily reflect the views
of the Office of Naval Research.
This work was also supported in part by the U.S. Department of
Transportation, Office of the Assistant Secretary for Research and Technology (USDOT/OST-R) under Grant No. 69A3551747126 through INSPIRE
University Transportation Center (http://inspire-utc.mst.edu) at Missouri
University of Science and Technology. The views, opinions, findings and
conclusions reflected in this publication are solely those of the authors and
do not represent the official policy or position of the USDOT/OST-R, or any
State or other entity.



\bibliographystyle{unsrt}
\bibliography{main}
\end{document}